\documentclass[letterpaper, 10 pt, conference]{IEEEtran}
\IEEEoverridecommandlockouts
\usepackage[dvipsnames]{xcolor}
\usepackage{graphicx} 
\usepackage{tikz}
\usetikzlibrary{shapes.callouts, shapes.geometric, shapes.symbols, arrows.meta, positioning, calc, backgrounds, fit}
\usepackage{lmodern} 
\usepackage{amsmath,amssymb,amsfonts}
\usepackage{hyperref}
\hypersetup{colorlinks,allcolors=black}
\usepackage[numbers]{natbib} 
\usepackage[most]{tcolorbox}
\usepackage{booktabs}
\usepackage{pifont}
\usepackage{makecell}
\usepackage[frozencache=true,cachedir=minted-cache]{minted}
\usepackage[caption=false,font=footnotesize]{subfig}
\usepackage{fontawesome5}
\usepackage{nicefrac} 
\usepackage{enumitem} 

\usepackage[all=normal, paragraphs=normal]{savetrees}
\usepackage{microtype} 

\usepackage[top=0.70in, bottom=0.75in, left=0.75in, right=0.75in]{geometry}

\renewcommand{\figurename}{Figure} 

\begin{document}

\title{\vspace*{0.15in} \LARGE \bf Adaptive Domain Modeling with Language Models:\\A Multi-Agent Approach to Task Planning}%

\author{Harisankar Babu$^{1,2}$, Philipp Schillinger$^{1}$, and Tamim Asfour$^{2}$%
\thanks{$^{1}$Bosch Center for Artificial Intelligence, Renningen, Germany.
        {\tt\footnotesize \{harisankar.babu, philipp.schillinger\}@bosch.com}}%
\thanks{$^{2}$Karlsruhe Institute of Technology, Karlsruhe, Germany.
        {\tt\footnotesize asfour@kit.edu}}%
}%

\maketitle

\bstctlcite{IEEEexample:BSTcontrol}

\begin{abstract}

We introduce TAPAS (Task-based Adaptation and Planning using AgentS), a multi-agent framework that integrates Large Language Models (LLMs) with symbolic planning to solve complex tasks without the need for manually defined environment models. TAPAS employs specialized LLM-based agents that collaboratively generate and adapt domain models, initial states, and goal specifications as needed using structured tool-calling mechanisms. Through this tool-based interaction, downstream agents can request modifications from upstream agents, enabling adaptation to novel attributes and constraints without manual domain redefinition. A ReAct (\textit{Reason+Act})-style execution agent, coupled with natural language plan translation, bridges the gap between dynamically generated plans and real-world robot capabilities. TAPAS demonstrates strong performance in benchmark planning domains and in the VirtualHome simulated real-world environment.

\end{abstract}

\renewcommand{\thefootnote}{\fnsymbol{footnote}}
\setlength{\textfloatsep}{6pt} 
\setlength{\parindent}{6pt}
\setlist[itemize]{noitemsep, nolistsep, leftmargin=15pt}
\setlist[enumerate]{noitemsep, nolistsep}

\section{Introduction}

Adaptability to unforeseen situations and evolving requirements is crucial for effective task planning in unstructured, open-world environments. For instance, dynamic environments require robots to handle incomplete information and changing task specifications. Large Language Models (LLMs) demonstrate strong generative capabilities, enabling the parsing of natural language into structured problem representations. However, leveraging these capabilities for automated planning remains an open challenge, especially in dynamic environments requiring contextual grounding, adaptability, and reasoning under uncertainty.

Classical symbolic planning frameworks, while powerful, face notable limitations. They rely on meticulously defined domain models that require significant human expertise. This makes them rigid and difficult to adapt to novel scenarios, thus limiting their applicability in dynamic, real-world settings. Current LLM-based approaches, conversely, lack the structured reasoning capabilities of symbolic planners. This work addresses the critical need for a planning system that combines the strengths of both: the adaptability of LLMs and the rigor of symbolic planning, thereby enhancing trustworthiness.

\begin{figure}[t]
    \centering
    \resizebox{3.25in}{!}{\begin{tikzpicture}[scale=1.0, every node/.style={font=\small}]


    \draw[thick] (-2.75,6) -- (2.75,6);

    \draw[fill=gray!20, thick] (-2.45,6) rectangle (-1.75,6.7);
    \node at (-2.1,6.35) {b1};

    \draw[fill=gray!40, thick] (-2.45,6.7) rectangle (-1.75,7.4);
    \node at (-2.1,7.05) {b2};

    \draw[fill=gray!60, thick] (-1.55,6) rectangle (-0.85,6.7);
    \node at (-1.2,6.35) {b3};

    \draw[fill=gray!20, thick, dashed] (0.85,6.7) rectangle (1.55,6.0);
    \node at (1.2,6.35) {b1};

    \draw[fill=gray!40, thick, dashed] (1.75,7.4) rectangle (2.45,6.7);
    \node at (2.1,7.05) {b2};

    \draw[fill=gray!60, thick, dashed] (1.75,6.7) rectangle (2.45,6.0);
    \node at (2.1,6.35) {b3};

    \draw[-Stealth, thick] (-0.5,6.7) -- (0.5,6.7);

    \node at (-3.75,6.5) {\Huge \faUser};  
    \node at (3.75,6.5) {\Huge \faRobot}; 

    \node[draw, rectangle callout, callout relative pointer={(0.3,-0.3)}, 
          align=center, text width=3cm, fill=white, font=\small] (bubbleTop) 
          at ($(-4.65,7.35)$)
          {Stack b2 on b3};

    \node[draw, rectangle callout, callout relative pointer={(-0.3,-0.3)}, 
          align=left, text width=4cm, fill=white, font=\small] (replyTop) 
          at ($(5.15,7.85)$)
          {\textcolor{blue}{\ding{224}} Adding goal: \texttt{(on b2 b3)} \\[0.5mm]
           \textcolor{blue}{\ding{224}} Solving... \\[0.5mm]
           \textcolor{green}{\ding{51}} Plan found! Executing...};


    \draw[thick] (-2.75,3) -- (2.75,3);

    \draw[fill=blue!20, thick] (-2.55,3) rectangle (-1.65,3.9);
    \node at (-2.1,3.45) {b1};

    \draw[fill=red!20, thick] (-1.45,3) rectangle (-0.75,3.7);
    \node at (-1.1,3.35) {b2};

    \draw[fill=green!20, thick] (-2.45,3.9) rectangle (-1.75,4.6);
    \node at (-2.1,4.25) {b3};

    \draw[fill=green!20, thick, dashed] (1.3,4.6) rectangle (2.0,5.3);
    \node at (1.65,4.95) {b3};

    \draw[fill=red!20, thick, dashed] (1.3,3.9) rectangle (2.0,4.6);
    \node at (1.65,4.25) {b2};

    \draw[fill=blue!20, thick, dashed] (1.2,3) rectangle (2.1,3.9);
    \node at (1.65,3.45) {b1};

    \draw[-Stealth, thick] (-0.45,3.7) -- (0.55,3.7);

    \node at (-3.75,3.0) {\Huge \faUser};  
    \node at (3.75,3.0) {\Huge \faRobot}; 

    \node[draw, rectangle callout, callout relative pointer={(0.2,-0.3)}, 
          align=center, text width=4cm, fill=white, font=\small] (bubbleBot) 
          at ($(-5.0,4.55)$)
          {Make a tower with the largest block on the bottom, the red block in the middle, and the green block on top.};

    \node[draw, rectangle callout, callout relative pointer={(-0.3,-0.3)}, 
          align=left, text width=6.0cm, fill=white, font=\small] (replyBot) 
          at ($(5.5,4.8)$)
          {\textcolor{red}{\ding{55}} Missing predicates: size, color. \\[0.5mm]
           \textcolor{orange}{\ding{45}} Updating domain:
           \begin{itemize}[leftmargin=20pt]
               \item Adding fluents: \texttt{size}, \texttt{color}.
               \item Updating stack action precondition.
           \end{itemize}
           \textcolor{blue}{\ding{224}} Adding goals...};

    \node at (-1.65,2.6) {\textbf{Initial State}};
    \node at (1.65,2.6) {\textbf{Goal State}};



\end{tikzpicture}}
    \caption{TAPAS dynamically adapts domain models to accommodate new goal constraints. If a goal can be represented with existing predicates, it directly generates and executes a plan. Otherwise, TAPAS detects missing predicates, updates the domain model by modifying action constraints, and integrates the new goal before solving. This allows symbolic planners to generalize beyond predefined representations while adapting to evolving task requirements.}
    \label{fig:yapa_illustration}
\end{figure}
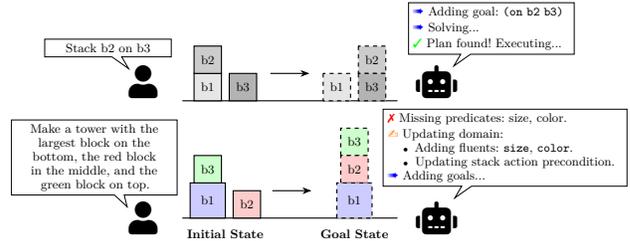

We introduce TAPAS (Task-based Adaptation and Planning using AgentS), a multi-agent framework that bridges natural language understanding and symbolic planning for adaptive task planning. Unlike traditional approaches, TAPAS employs a set of specialized LLM-based agents, each responsible for a distinct phase of problem formulation: domain modeling, initial state generation, and goal specification. This modular design ensures structured problem representation while allowing flexible adaptation to new task constraints. Using the Unified Planning\footnotemark[1] (UP) framework, TAPAS adaptively updates domain representations through iterative refinement, automatically incorporating novel attributes such as object properties or action constraints. Figure~\ref{fig:yapa_illustration} illustrates how TAPAS dynamically adapts domain models to new goal constraints.

\footnotetext[1]{The AIPlan4EU Unified Planning Library: \url{https://github.com/aiplan4eu/unified-planning}}

Key contributions of TAPAS include:
\begin{enumerate}
    \item An approach to reason over and dynamically adapt to unexpected goal specifications, including those that introduce novel attributes or constraints, without requiring manual domain redefinition. This adaptation is achieved through a modular, collaborative framework where LLM agents autonomously generate and update structured domain models, initial states, and goals.
    \item A robust planning and execution pipeline that bridges the gap between dynamically generated symbolic plans and real-world robot capabilities. This is achieved through natural language translation of plans, ReAct (\textit{Reason+Act})-based execution to handle domain-skill discrepancies, and iterative feedback-driven validation.
\end{enumerate}

By leveraging these features, TAPAS provides a scalable and adaptive approach to automated planning, addressing traditional barriers in symbolic planning. Its modular agent-based design and feedback-driven execution make it well suited for open-world environments where contextual reasoning and adaptability are essential.


\section{Related Work}

Effective task planning in dynamic, open-world environments requires reasoning and adaptability, which traditional methods often lack. Recently, LLMs have shown remarkable progress in reasoning, offering new avenues for these challenges. Foundational prompting techniques, such as Chain of Thought (CoT)~\cite{wei2022chain, kojima2022large}, established that LLMs can perform in-context reasoning. Subsequent methods like Tree of Thoughts (ToT)~\cite{shunyu2024tree} and Graph of Thoughts (GoT)~\cite{besta2024graph} further enhanced problem-solving by introducing more structured exploration of the solution space.

Building on these advancements, LLMs have been directly applied to task planning as planners. Early approaches explored translating high-level language into actionable steps~\cite{huang22a} and grounding plans in robotic affordances~\cite{ichter23a}. To improve robustness, subsequent methods introduced iterative self-refinement through feedback loops~\cite{huang23c, shinn23} and tool-interactive correction~\cite{2024critic}. \textit{ReAct}~\cite{2023react} enables interleaved reasoning and action execution, improving adaptability in dynamic tasks. Other works have focused on grounding plans in 3D scene representations~\cite{rana23a}, integrating strategic look-ahead with Monte Carlo Tree Search~\cite{hao2023rap}, and assessing generalization of LLM-based planners across multiple domains~\cite{silver2024}. However, PlanBench~\cite{valmeekam2023planbench} benchmarked LLM planning capabilities, highlighting gaps compared to traditional symbolic planners, motivating hybrid neuro-symbolic frameworks where LLMs assist, rather than replace, symbolic methods~\cite{kambhampati2024}.

Consequently, hybrid models that integrate LLMs with symbolic planners have gained attention. \textit{LLM+P}~\cite{liu2023llm+} combines Planning Domain Definition Language (PDDL)-based symbolic planners with LLMs to translate natural language descriptions into structured problem definitions, while \textit{LLM-DP}~\cite{dagan2023} demonstrates how LLMs complement symbolic planners in handling uncertainty in embodied tasks. Similarly, \citet{worldmodels2023} and \textit{NL2Plan}~\cite{nl2plan} use multistep frameworks to generate and refine PDDL models and have been extended to multi-agent, long-horizon tasks~\cite{lamma-p}. Other approaches couple LLMs with alternative formalisms, such as Answer Set Programming (ASP)~\cite{clmasp, yang2023} or signal temporal logic~\cite{autotamp}, to enhance robust reasoning for robotic planning.

TAPAS builds on these hybrid approaches, particularly those that use LLMs for problem formulation~\cite{liu2023llm+, birr24, liu2024delta}. However, unlike systems that assume a static problem formulation, TAPAS enables iterative domain adaptation. Its agents can dynamically detect missing constraints, modify domain fluents, and refine symbolic representations.

Furthermore, TAPAS integrates a procedural memory for skill transfer~\cite{pohl2024, alimem} with its dynamic domain adaptation capability to continuously learn from novel problem structures.


\section{Adaptive Domain Modeling via LLM Agents}

TAPAS's ability to handle changing environments and novel task specifications centers on its adaptive domain modeling capability. This section details this core mechanism by describing the multi-agent framework and the internal architecture of the LLM agents that collaboratively construct and adapt planning domains.

\subsection{Multi-Agent Planning Framework}

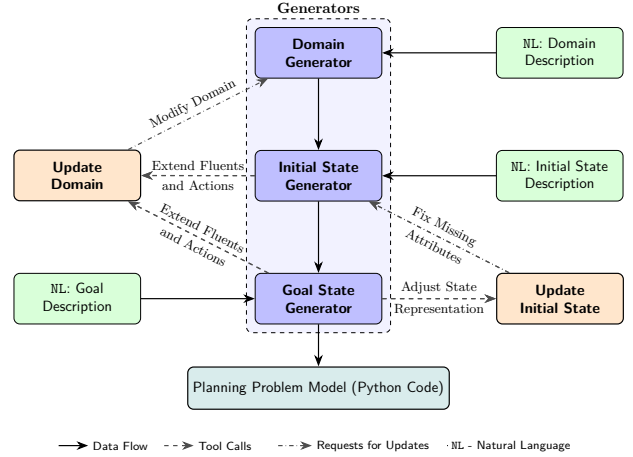
\begin{figure}[!t]
    \centering
    \resizebox{3.25in}{!}{
\tikzset{
natural/.style={draw, fill=green!15, rounded corners, minimum width=3cm, minimum height=1.2cm, font=\sffamily},
generator/.style={draw, fill=blue!25, rounded corners, minimum width=3cm, minimum height=1.2cm, font=\sffamily\bfseries},
output/.style={draw, fill=teal!15, rounded corners, minimum width=3.5cm, minimum height=1.0cm, font=\sffamily, thick},
update/.style={draw, fill=orange!20, rounded corners, minimum width=3cm, minimum height=1.2cm, font=\sffamily\bfseries, draw=black, thick},
group/.style={draw, dashed, rectangle, rounded corners, inner sep=0.2cm, label={[font=\bfseries]above:#1}, fill=blue!5},
legend/.style={font=\sffamily\footnotesize},
arrow/.style={-{Stealth[scale=1.2]}, thick},
dashdotarrow/.style={dash dot, -{Stealth[scale=1.2]}, thick, draw=black!70}, 
dashedarrow/.style={dashed, -{Stealth[scale=1.2]}, thick, draw=black!70},
textarrow/.style={midway, sloped, font=\small, above},
textarrow_b/.style={midway, sloped, font=\small, below},
}

\begin{tikzpicture}[node distance=1.7cm and 2.7cm,
        every node/.style={align=center, font=\sffamily}]

    \node[generator] (domainGen) {Domain\\Generator};
    \node[generator, below=of domainGen] (initGen) {Initial State\\Generator};
    \node[generator, below=of initGen] (goalGen) {Goal State\\Generator};

    \begin{scope}[on background layer]
        \node[group=Generators, fill=blue!5] (generatorGroup) [fit=(domainGen) (initGen) (goalGen)] {};
    \end{scope}

    \node[natural, right=of domainGen] (domainDesc) {\texttt{NL}: Domain\\Description};
    \node[natural, right=of initGen] (initDesc) {\texttt{NL}: Initial State\\Description};
    \node[natural, left=of goalGen] (goalDesc) {\texttt{NL}: Goal\\Description};

    \node[update, above=of goalDesc] (domainUpd) {Update\\Domain};
    \node[update, right=of goalGen] (iniUpd) {Update\\Initial State};

    \node[output, below=1.0cm of goalGen] (planningProblem) {Planning Problem Model (Python Code)};

    \draw[arrow] (domainDesc) -- (domainGen);
    \draw[arrow] (domainGen) -- (initGen);
    \draw[arrow] (initGen) -- (goalGen);
    \draw[arrow] (goalDesc) -- (goalGen);
    \draw[arrow] (initDesc) -- (initGen);
    \draw[arrow] (goalGen) -- (planningProblem);

    \draw[dashedarrow] (initGen) -- (domainUpd) node[textarrow] {Extend Fluents};
    \draw[dashedarrow] (initGen) -- (domainUpd) node[textarrow_b] {and Actions};
    \draw[dashedarrow] (goalGen) -- (domainUpd) node[textarrow] {Extend Fluents};
    \draw[dashedarrow] (goalGen) -- (domainUpd) node[textarrow_b] {and Actions};
    \draw[dashdotarrow] (domainUpd) -- (domainGen) node[textarrow] {Modify Domain};
    \draw[dashedarrow] (goalGen) -- (iniUpd) node[textarrow] {Adjust State};
    \draw[dashedarrow] (goalGen) -- (iniUpd) node[textarrow_b] {Representation};
    \draw[dashdotarrow] (iniUpd) -- (initGen) node[textarrow] {Fix Missing};
    \draw[dashdotarrow] (iniUpd) -- (initGen) node[textarrow_b] {Attributes};

    \node[legend, below=0.5cm of planningProblem] {
        \begin{tikzpicture}
        \draw[arrow] (0,0) -- (0.7,0) node[right] {Data Flow};
        \draw[dashedarrow] (2.5,0) -- (3.2,0) node[right] {Tool Calls};
        \draw[dashdotarrow] (5.2,0) -- (6.0,0) node[right] {Requests for Updates};
        \draw[] (9.2,0) -- (9.2,0) node[right] {\texttt{NL} - Natural Language};
        \end{tikzpicture}
    };

\end{tikzpicture}}
    \caption{Multi-LLM-agent planning framework overview. The \textit{Domain}, \textit{Initial State}, and \textit{Goal State} Generators iteratively refine the planning problem. When a goal requires fluents not yet represented, the system detects missing attributes and requests domain model updates, ensuring coherent problem formulation. The final output is a planning problem in Python.}
    \label{fig:yapa-architecture-1}
\end{figure}

TAPAS employs a multi-agent planning framework integrating the generative capabilities of LLMs with the structured reasoning of symbolic planning. The framework consists of three specialized LLM-based agents: the \textit{Domain Generator}, \textit{Initial State Generator}, and \textit{Goal State Generator}. These agents operate autonomously, but collaboratively to construct well-defined planning problems. By leveraging both independent reasoning and inter-agent interactions, TAPAS effectively handles complex and dynamic environments. Each agent's system prompt emphasizes tool use and discourages assumptions.

The architecture of TAPAS's multi-agent planning framework is illustrated in Figure~\ref{fig:yapa-architecture-1}. The framework operates as follows:

\begin{itemize}
    \item \textit{Domain Generator}: Receives a natural language problem description and converts it into domain code, defining types, fluents, and actions in Python.
    \item \textit{Initial State Generator}: Receives the domain code and a natural language description of the initial state. Generate the initial state code specifying object instances and their initial fluent values.
    \item \textit{Goal State Generator}: Receives the domain code, the initial state code, and a natural language description of the goal. Defines the goal conditions, ensuring consistency with the initial state and the domain model.
\end{itemize}

Each agent validates its output both semantically and during runtime. If an agent detects errors, it iteratively refines its output until a valid solution is achieved or a predefined correction limit is reached.

\textit{Collaborative Refinement and Adaptation via Tool Use:} A key strength of TAPAS is its iterative refinement process, in which LLM-based agents actively detect, correct, and adapt planning problem definitions in response to evolving constraints through structured tool use. Downstream generators analyze upstream output, detecting missing attributes or ill-defined elements. When inconsistencies arise, agents engage in a structured feedback loop, invoking specific tools to request modifications, as detailed in Figure~\ref{fig:agent_tools}.

\begin{figure}[tb]
    \centering
    \begin{tcolorbox}[width=\linewidth, colframe=blue!50!black, colback=blue!2, boxrule=0.5pt, arc=2pt, left=3pt, right=3pt, top=1.5pt, bottom=1.5pt, fontupper=\small, fontlower=\small, fonttitle=\small]
        \textbf{Agent Tools for Domain Modification:}
        \begin{itemize}
            \item Tool name: \texttt{missing\_or\_incorrect\_fluent}\\
            Arguments: \texttt{fluent\_name}, \texttt{fluent\_description}\\ \textit{Request the addition or modification of a fluent.}
            \item Tool name: \texttt{action\_modification}\\
            Arguments: \texttt{action\_name}, \texttt{change\_description}\\
            \textit{Requests modifications to an existing action.}
        \end{itemize}
        \tcblower
        \textbf{Agent Tools for Initial State Modification:}
        \begin{itemize}
            \item Tool name: \texttt{missing\_objects}\\
            Arguments: \texttt{object\_type}, \texttt{object\_description}\\ \textit{Requests the addition of objects required for the goal.}
        \end{itemize}
    \end{tcolorbox}
    \vspace*{-0.8em}
    \caption{Agent tools for modifying domain and initial state.}
    \label{fig:agent_tools}
\end{figure}

The Initial State Generator may also query the user for additional information (e.g., the color of an object) if not provided in the initial description. This tool-based interaction ensures that only downstream agents trigger changes in upstream components, maintaining a consistent and well-defined problem structure. The tools are modular, allowing for easy extension with new capabilities.

For instance, in the classic \textit{blocksworld} domain, blocks are typically represented as objects (e.g., \textit{b1, b2, b3}). If the user specifies a goal such as \textit{``place the blue block on top of the red block"}, the \textit{Goal State Generator} identifies that color attributes are absent from the initial state and domain model. It then invokes the \texttt{missing\_or\_incorrect\_fluent} tool to request adding a \texttt{color} fluent. This, in turn, prompts the \textit{Initial State Generator} to incorporate these attributes and the \textit{Domain Generator} to update the domain model. This bottom-up, tool-mediated adaptation ensures consistency across all problem levels. This mechanism allows TAPAS to handle a wide range of novel constraints and attributes, not just predefined ones.

By iteratively resolving inconsistencies through structured tool use, TAPAS surpasses traditional planning approaches that rely on static problem formulations. Agents not only correct errors but also evolve problem representation in response to new constraints or goals, enabling flexible and adaptive planning in dynamic real-world environments.

\subsection{Agent Architecture}

Each agent in TAPAS’s modular architecture integrates reasoning, refinement, and memory. Agents operate independently but communicate via structured tool-calls, ensuring coherent problem formulation and enabling flexible task decomposition and adaptation.

A TAPAS agent's workflow comprises three key phases: \textit{Generation}, where the agent produces structured representations of planning problems based on input queries and available context; \textit{Evaluation and Refinement}, where the agent assesses its output, incorporates relevant feedback, and refines the response if necessary; and \textit{Execution and Feedback}, where the validated output is executed within the planning pipeline, with any execution feedback integrated into the reasoning process.

\subsubsection{Memory Mechanisms}

To support task adaptation, TAPAS primarily relies on a short-term context memory ($\mathcal{M}_{\text{context}}$), a buffer which stores recent interactions like user queries, tool calls, and agent responses. This enables multi-turn coherence and keeps the agent focused on the immediate task. As an auxiliary mechanism for long-term learning across different tasks, TAPAS also includes a procedural memory ($\mathcal{M}_{\text{procedural}}$), a long-term repository for storing generalizable corrections.

\textit{Memory Storage and Retrieval:} The procedural memory is designed to learn incrementally from individual, significant events. Storage is prompted by corrections deemed to be generalizable, which are typically those resulting from explicit user feedback. Upon identifying such a correction, the agent invokes a dedicated memory storage tool. For retrieval, given a new task query $q$, the agent computes similarity scores between $q$ and stored summaries $\{s_i\}$ in $\mathcal{M}_{\text{procedural}}$: ${\text{score}(q, s_i) = \nicefrac{e_q \cdot e_{s_i}}{\|e_q\| \|e_{s_i}\|}}$, where $e_q$ and $e_{s_i}$ are embedding representations of the task query and stored feedback summary, respectively. High-scoring entries are integrated into the agent's reasoning process. However, this similarity-based retrieval has limitations, as it can retrieve memories that are syntactically similar but contextually irrelevant, potentially leading the agent to generate a flawed or unsolvable planning problem.

\subsubsection{Self-Reflection Mechanism}
To enhance response quality, TAPAS employs a self-reflection mechanism where a generator agent collaborates with an LLM critic. The critic evaluates the generator's initial response, $r$, providing detailed feedback and a quality score $\sigma \in [0,1]$. Instructions for the critic to determine $\sigma$ include criteria like correctness, coherence, and completeness, as well as a quality threshold $\tau$ for acceptance.
The critic acts as an evaluator, analyzing the response and generating both a numerical score and textual feedback explaining any identified issues or areas for improvement.

If $\sigma \geq \tau$, the response is deemed acceptable. Otherwise, the generator incorporates the critic's feedback to refine its response. This iterative process continues until the score meets the threshold or a predefined iteration limit is reached. If the limit is reached, the system proceeds with the latest response. This represents a design trade-off between robustness and computational cost, as accepting a lower-quality response may lead to subsequent errors in downstream modules or a plan failure.

\section{Plan Solving and Execution}

\begin{figure}[!t]
    \centering
    \resizebox{3.25in}{!}{\tikzset{
    generator/.style={draw, fill=blue!30, rounded corners, minimum width=3.2cm, minimum height=1.2cm, font=\sffamily},
    generator2/.style={draw, dashed, thick, fill=blue!30, rounded corners, minimum width=3.2cm, minimum height=1.2cm, font=\sffamily},
    output/.style={draw, fill=teal!20, rounded corners, minimum width=3.2cm, minimum height=1.2cm, font=\sffamily},
    solver/.style={draw, fill=red!20, rounded corners, minimum width=3.2cm, minimum height=1.2cm, font=\sffamily},
    external1/.style={draw, ellipse, fill=lime!20, minimum width=2.8cm, minimum height=1.2cm, font=\sffamily},
    external2/.style={draw, ellipse, fill=orange!20, minimum width=2.8cm, minimum height=1.2cm, font=\sffamily},
    group/.style={draw, dashed, thick, rounded corners, inner sep=0.2cm, label={[font=\bfseries]below:#1}, fill=blue!5},
    group2/.style={draw, dashed, thick, rounded corners, inner sep=0.2cm, label={[font=\bfseries]below:#1}, fill=yellow!10},
    arrow/.style={-{Stealth[scale=1.2]}, thick},
    dashedarrow/.style={dashed, -{Stealth[scale=1.2]}, thick},
    textarrow/.style={midway, font=\small},
}

    \begin{tikzpicture}[node distance=1.2cm and 2.0cm, every node/.style={align=center, font=\sffamily}]
    
        \node[output] (planningProblem) {Planning Problem\\Model};
        \node[generator2, fill=red!20, below=of planningProblem] (solverMain) {Solver-Debugger};
        \node[generator, below=of solverMain] (translator) {Abstraction};
        \node[generator2, below=of translator] (executorMain) {Plan Executor};
        \node[output, below=of executorMain] (executionEnd) {Execution\\Complete/Failure};
    
        \draw[arrow] (planningProblem) -- (solverMain);
        \draw[arrow] (solverMain) -- (translator) node[textarrow, right] {Structured Plan};
        \draw[arrow] (translator) -- (executorMain) node[textarrow, right] {Plan in NL};
        \draw[arrow] (executorMain) -- (executionEnd);
    
        \node[solver, right=1.5cm of solverMain] (solver) {Solver};
        \node[generator, right=2.5cm of solver] (debugger) {Debugger};
    
        \draw[arrow] (solver.10) -- (debugger.170) node[textarrow, above] {Report Errors};
        \draw[arrow] (debugger.190) -- (solver.350) node[textarrow, below] {Provide Fixes};
    
        \begin{scope}[on background layer]
            \node[group=Solver-Debugger, fill=red!5] (solverGroup) [fit=(solver) (debugger)] {};
        \end{scope}
    
        \node[generator, right=1.5cm of translator, yshift=-1.25cm] (actionExecutor) {Action Executor};
        \node[generator, below=of actionExecutor] (validator) {Validator};
    
        \begin{scope}[on background layer]
            \node[group=Plan Executor, fill=blue!5] (executorGroup) [fit=(actionExecutor) (validator)] {};
        \end{scope}
    
        \node[external2, right=2.75cm of executorGroup] (sensors) {Sensors};
        \node[external1, above=0.3cm of sensors] (control) {Controllers};
        \begin{scope}[on background layer]
            \node[group2=Robot, fill=yellow!10] (robot) [fit=(sensors) (control)] {};
        \end{scope}
    
        \draw[dashedarrow] (actionExecutor.10) -- (control) node[textarrow, above] {API Calls};
        \draw[dashedarrow] (sensors.west) -- ++(-0.5,0) |- (actionExecutor.east);
        \draw[dashedarrow] (sensors.west) -- ++(-0.5,0) |- (validator.east)  node[textarrow, below, pos=0.70] {Sensor Data};
    
        \draw[arrow] (actionExecutor.250) -- (validator.110) node[textarrow, left] {Execution\\End};
        \draw[arrow] (validator.70) -- (actionExecutor.290) node[textarrow, right] {Execution\\Feedback};

        \draw[dashed, thick, gray] (solverMain.18) -- (solverGroup.171);
        \draw[dashed, thick, gray] (solverMain.342) -- (solverGroup.189);
        \draw[dashed, thick, gray] (executorMain.north east) -- (executorGroup.135);
        \draw[dashed, thick, gray] (executorMain.south east) -- (executorGroup.225);
    
    \end{tikzpicture}}
    \caption{TAPAS's execution pipeline. The solver generates a structured plan, translated into natural language instructions and executed by the plan executor. The plan executor consists of the action executor, which performs actions, and the validator, which monitors execution and determines whether to continue or terminate based on goal fulfillment.}
    \label{fig:yapa-architecture-2}
\end{figure}
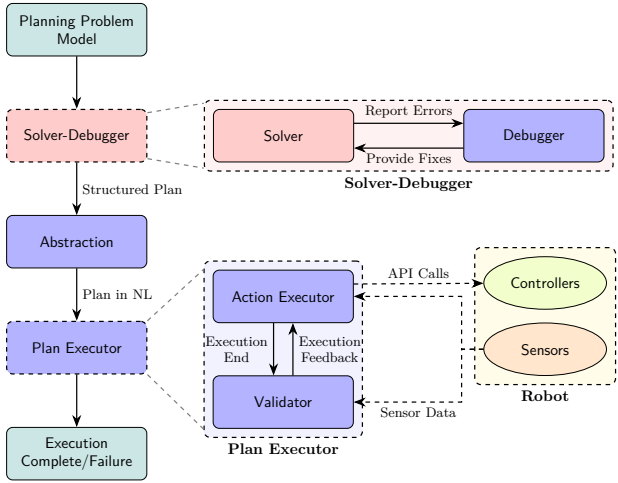

Once the planning problem is defined, the next challenge is generating and executing a plan. This section describes the components for plan generation and plan execution. Figure~\ref{fig:yapa-architecture-2} overviews this pipeline. A key TAPAS feature is its ability to handle generated domain models, where action names and arguments may not directly correspond to available robot skills, requiring the flexible plan solving and execution approach described below.

\subsection{Solver}

The solver uses the UP framework, which supports multiple underlying planning formalisms like PDDL, to generate a sequential plan from the provided domain, initial state, and goal specifications. If semantic or modeling errors are detected, the solver invokes a Retrieval Augmented Generation (RAG)-assisted LLM debugger.

The debugger queries a vector database containing UP framework documentation. The database is constructed offline by embedding relevant UP framework documentation using a pre-trained sentence embedding model. This process provides concrete code corrections to resolve solver errors, improving robustness and helping to ensure the generated plan adheres to problem constraints.

\subsection{Plan Abstraction}

Because the domain model is generated, its actions might not have a one-to-one correspondence with the robot's available skills; they might have different names and arguments. Therefore, direct execution of the solver's output might be impossible. The purpose of plan abstraction is to bridge this semantic gap by decoupling the high-level symbolic plan from the robot’s specific, low-level skill library.

\begin{figure}[tb]
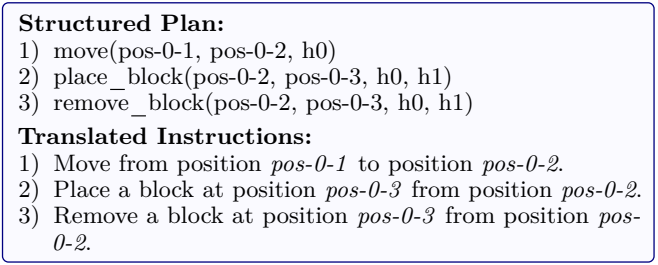

    \centering
    \begin{tcolorbox}[width=\linewidth, fontupper=\small, fontlower=\small, colframe=blue!50!black, colback=blue!2, boxrule=0.5pt, arc=2pt, left=3pt, right=3pt, top=1.5pt, bottom=1.5pt]
        \textbf{Structured Plan:}
        \begin{enumerate}[leftmargin=*]
            \item move(pos-0-1, pos-0-2, h0)
            \item place\_block(pos-0-2, pos-0-3, h0, h1)
            \item remove\_block(pos-0-2, pos-0-3, h0, h1)
        \end{enumerate}
        \vspace{0.3em}
        \textbf{Translated Instructions:}
        \begin{enumerate}[leftmargin=*]
            \item Move from position \textit{pos-0-1} to position \textit{pos-0-2}.
            \item Place a block at position \textit{pos-0-3} from position \textit{pos-0-2}.
            \item Remove a block at position \textit{pos-0-3} from position \textit{pos-0-2}.
        \end{enumerate}
    \end{tcolorbox}
    \vspace*{-0.8em}
    \caption{A structured plan translated into natural language.}
    \label{fig:plan_translation}
\end{figure}

To achieve this, the LLM-based plan abstraction converts the solver's structured plan into natural language instructions. This translation serves as an intermediate representation that facilitates the subsequent grounding of abstract plan steps (e.g., ``place a block") to concrete skills by a dedicated execution agent, as described in the following section. This process makes the plan understandable to human operators and is also suitable for modern vision-language-action (VLA) models~\cite{openvla, openxemb}. An example of a structured plan and its translation is shown in Figure~\ref{fig:plan_translation}.

Here, internal parameters such as height (\textit{h0}, \textit{h1}) are omitted since they do not affect fundamental task understanding. This ensures the translated instructions are concise, user-friendly, and actionable. Despite this translation, the generated plan might not be directly executable and requires further processing by a dedicated executor.

\subsection{Plan Executor}

To perform the crucial step of aligning the abstracted natural language plan with the robot's available skills, the plan executor employs a flexible strategy within the execution environment. It consists of two LLM-based agents: the \textit{Action Executor Agent} and the \textit{Validator Agent}, which collaborate to ensure successful execution.

The core execution workflow is as follows:

\begin{enumerate}
    \item The executor receives the current environment state. This can be text-based (e.g., symbolic state descriptions) or combine text and images, providing a view of the environment along with textual state descriptions.
    \item The Action Executor Agent sequentially processes each step in the translated plan, treating each as a short-term sub-goal. Action execution uses a ReAct-style approach~\cite{2023react} to map the natural language instruction to the most appropriate available skill (tool) and its corresponding parameters.
    \item The Action Executor Agent then executes the selected skill.
\end{enumerate}

After the Action Executor Agent completes the action sequence or encounters an unrecoverable error, the Validator Agent reviews the execution logs and the final environment state. If the overall goal is met, the process terminates. Otherwise, the Validator provides corrective feedback to re-invoke the Action Executor with modified instructions, potentially attempting a different skill sequence. For critical failures, the Validator notifies the user instead of continuing execution.


\begin{table*}[t]
        \centering
        \caption{Accuracy of TAPAS on LLM+P benchmark domains.}
        \label{tab:planning-results}
        \renewcommand{\arraystretch}{1.0}
        \begin{tabular}{l c c c c c c c}
                \specialrule{1.2pt}{0pt}{0pt}
                \noalign{\vspace{1mm}}
                \textbf{}            & \textbf{}                      & \multicolumn{5}{c}{\textbf{Problems Solved (\%)}} & \textbf{}                       \\
                \noalign{\vspace{-1mm}}
                \multicolumn{2}{l}{} & \multicolumn{5}{c}{\hrulefill} &                                                                                     \\
                \shortstack{\textbf{Domain}                                                                                                                 \\ \textbf{Name}} & \shortstack{\textbf{Generation} \\ \textbf{Status}} & GPT-4o & \shortstack{Claude 3.7 \\ Sonnet} & \shortstack{GPT-4o \\ Mini} & \shortstack{Mistral \\ Large} & \shortstack{Cohere \\ command-r} & \shortstack{\textbf{LLM+P} \\ \textbf{Reported (\%)}} \\
                \hline
                Barman               & \ding{51}                      & 97                                                & \textbf{100}       & 10 & 15 & 25            & 20  \\
                Blocksworld          & \ding{51}                      & \textbf{100}                                      & 95                 & 70 & 5  & \textbf{100}  & 90  \\
                Floortile            & \ding{51}                      & 57                                                & \textbf{100}       & 0  & 15 & 0             & 0   \\
                Grippers             & \ding{51}                      & \textbf{100}                                      & \textbf{100}       & 50 & 15 & \textbf{100}  & \textbf{100} \\
                Storage              & \ding{51}                      & 90                                                & \textbf{100}       & 0  & 20 & 20            & 85  \\
                Termes               & \ding{51}                      & 95                                                & \textbf{100}       & 0  & 15 & 0             & 20  \\
                Tyreworld            & \ding{51}                      & 80                                                & \textbf{95}        & 0  & 0  & 0             & 90  \\
                \specialrule{1.2pt}{0pt}{0pt}
        \end{tabular}
\end{table*}

\section{Results}

We evaluated TAPAS on benchmark planning problems from LLM+P~\cite{liu2023llm+} and additional experiments designed to assess its ability to adaptively update domain representations and adapt to new constraints. The experiments focus on three key aspects: (1) domain generation and planning accuracy on standard benchmarks, (2) handling new attributes in goal specification, and (3) evaluating the framework in the \textit{VirtualHome} simulation environment~\cite{puig2018virtualhome}.

\paragraph{Experimental Setup}
All experiments utilized the GPT-4o\footnotemark[2] model (version 2024-08-06) via the Azure OpenAI Service API. The model was used without fine-tuning, with a single generic example of the UP framework provided in the system prompt to avoid in-domain influence. A temperature of 0.0 was used unless otherwise specified, with default values for \textit{top-p} and other parameters. To prevent infinite loops, a recursion limit of 10 was enforced, restricting the maximum number of iterative steps in the planning process.

\footnotetext[2]{GPT-4o: \url{https://cdn.openai.com/gpt-4o-system-card.pdf}}

\subsection{Benchmark Evaluation on LLM+P Domains}

TAPAS was evaluated on seven planning domains from LLM+P~\cite{liu2023llm+}, encompassing a range of classical planning tasks. The evaluation consisted of two stages: domain generation and problem solving using the generated domain. Note that LLM+P utilized GPT-4 for its evaluation along with in-domain example.

For domain generation, TAPAS inferred types, fluents, and action models from natural language descriptions  of each domain, including actions, preconditions, and effects. Downstream agents provided feedback for modifications in case of inconsistencies or missing fluents. The generated domains were then used with a planner, and the solutions were validated against ground-truth data. Context memory was cleared between runs to ensure independence and no procedural memory was used.

\paragraph{Results on Domain Generation}
TAPAS successfully generated executable domain models for all seven domains. Downstream agent feedback was minimal, primarily limited to the \textit{barman} domain, where the initial state generator identified missing fluents related to cocktail ingredients, leading to a refinement of the model.

\paragraph{Results on Planning}
Each generated domain was used to solve all 20 problem instances from the corresponding benchmark dataset.
To test the robustness of the downstream agents, the successfully generated domain for each benchmark was provided directly, bypassing the initial domain generation stage. The initial and goal state generators were still permitted to request modifications, which assessed their ability to work with a correct domain without introducing flawed changes.

Table~\ref{tab:planning-results} presents TAPAS's accuracy across the domains, calculated as the average percentage of correctly solved problems over two repetitions of the 20 problems. The model consistently produced correct solutions, demonstrating strong generalization across diverse planning tasks.

TAPAS demonstrated robustness against false positive domain modification requests from downstream agents. When fluents were mistakenly reported as missing, the domain generator correctly rejected unnecessary changes, maintaining the original model. Most failures occurred in the \textit{floortile} and \textit{tyreworld} domains. In \textit{floortile}, the LLM misidentified directional fluents, resulting in incorrect movement constraints. In \textit{tyreworld}, additional fluents were introduced, leading to over-specified goals. Notably, these errors were not correlated with problem complexity, such as the number of objects, but with the interpretation of domain constraints.

\begin{table}[tb]
        \centering
        \caption{Impact of temperature on planning success rate}
        \label{tab:planning-results-temp}
        \renewcommand{\arraystretch}{1.2}
        \begin{tabular}{l c c c}
                \specialrule{1.2pt}{0pt}{0pt}
                \noalign{\vspace{0.8mm}}
                \textbf{}            & \multicolumn{3}{c}{\textbf{Temperature}}                        \\
                \noalign{\vspace{-1.4mm}}
                \multicolumn{1}{l}{} & \multicolumn{3}{c}{\hrulefill}                                  \\
                \textbf{Domain}      & 0.0                                      & 0.1                     & 0.3                 \\
                \hline
                Barman               & 97                                       & \textbf{100}            & \textbf{100}   \\
                Blocksworld          & \textbf{100}                             & \textbf{100}            & 95                  \\
                Floortile            & 57                                       & \textbf{70}             & 47                  \\
                Grippers             & \textbf{100}                             & \textbf{100}            & 97                 \\
                Storage              & 90                                       & \textbf{92}             & 82                  \\
                Termes               & 95                                       & \textbf{100}            & 97                  \\
                Tyreworld            & 80                                       & \textbf{85}             & 67                  \\
                \hline
                \textbf{Average}     &
                \textbf{88.42}       & \textbf{92.42}                           & \textbf{83.57}       \\
                \specialrule{1.2pt}{0pt}{0pt}
        \end{tabular}
\end{table}

To investigate factors influencing TAPAS’s performance, ablations were performed using different closed and open-source models and temperature settings. Model ablations were performed using Claude 3.7 Sonnet\footnotemark[3], GPT-4o Mini, Mistral Large\footnotemark[4] - 2411, and Cohere Command-R\footnotemark[5] - 08-2024, across all previously generated domains from GPT-4o. Additionally, the impact of sampling temperatures between 0.0 and 0.3 was evaluated, as higher values can introduce variability in code generation~\cite{renze-2024-effect}. Table~\ref{tab:planning-results-temp} displays the average observed success rate for each temperature in multiple runs. Performance decreased at a temperature of 0.3, as LLM introduced extraneous fluents and deviated from system instructions. However, this behavior may be advantageous in scenarios requiring adaptive extension of domain attributes.

\footnotetext[3]{Claude 3.7: \url{https://www.anthropic.com/news/claude-3-7-sonnet}}
\footnotetext[4]{Mistral Large 2: \url{https://mistral.ai/news/mistral-large-2407}}
\footnotetext[5]{Command R: \url{https://docs.cohere.com/v2/docs/command-r}}


\subsection{Evaluation on New Attributes in Goal Specification}

To evaluate TAPAS's adaptability to novel goal constraints, we conducted experiments requiring domain modifications that would typically need manual extension. The tests involved introducing \texttt{color} and \texttt{size} attributes to \textit{blocksworld}, and numeric \texttt{battery\_level} constraints to the \textit{grippers} and \textit{floortile} domains. The results in four sets of five problems are summarized in Table~\ref{tab:adaptability-results}.

\begin{table}[tb]
    \centering
    \caption{Adaptability to Novel Goal Constraints}
    \label{tab:adaptability-results}
    \renewcommand{\arraystretch}{1.1}
    \begin{tabular}{l l c}
        \specialrule{1.2pt}{0pt}{0pt}
        \noalign{\vspace{0.8mm}}
        \textbf{Domain}      & \textbf{New Constraint Type} & \textbf{Success Rate} \\
        \hline
        Blocksworld & Color-based goal               & 100\%         \\
        Blocksworld & Size-based goal and action     & 90\%          \\
        Grippers    & Battery level goal and action  & 100\%         \\
        Floortile   & Battery level goal and action  & 70\%          \\
        \specialrule{1.2pt}{0pt}{0pt}
    \end{tabular}
\end{table}

When a goal required a new descriptive attribute, such as color, TAPAS correctly inferred the need for a new fluent. The \textit{Domain Generator} added a \texttt{color} predicate to the model, and the \textit{Initial State Generator} then queried the user for the color of each block, ensuring a consistent and solvable problem definition.

More significantly, TAPAS demonstrated the ability to modify core action logic to satisfy functional constraints. For goals involving object size or battery consumption, the system updated action preconditions and effects. For instance, to handle a size-based constraint, TAPAS introduced a \texttt{size} fluent and altered the \texttt{stack} action.
This required updating the action's precondition to enforce the new stacking rule as shown in Figure~\ref{fig:adaptability-diff}.

\begin{figure}[tb]
    \centering
    \begin{tcolorbox}[width=\linewidth, fontupper=\small, colframe=blue!50!black, colback=blue!2, boxrule=0.5pt, arc=2pt, left=3pt, right=3pt, top=1.5pt, bottom=1.5pt]
    \textbf{New Goal:} \textit{``The goal is to move the blocks to make a tower with the largest block on the bottom and the smallest block on top. Ensure that a block can be stacked only on top of a larger block in the action.''}
    \end{tcolorbox}

    \definecolor{added}{rgb}{0.0, 0.5, 0.0}
    \definecolor{deleted}{rgb}{0.6, 0.0, 0.0}
    
    \begin{tcolorbox}[width=\linewidth, colframe=lime!50!black, colback=lime!0, boxrule=0.5pt, arc=2pt, left=3pt, right=3pt, top=1.5pt, bottom=1.5pt]
    \begin{minted}[
    fontsize=\small,
    breaklines=true,
    breakanywhere=true,
    escapeinside=||,
    style=tango
    ]{diff}
    |\textcolor{deleted}{- (:requirements :strips :typing)}|
    |\textcolor{added}{+ (:requirements :strips :typing :numeric-fluents)}|
      (:types
        block - object
      )
    |\textcolor{added}{+ (:functions (size ?b))}|
      (:action stack
      :parameters (?b1 ?b2 - block)
    |\textcolor{deleted}{- :precondition (and (holding ?b1) (clear ?b2))}|
    |\textcolor{added}{+ :precondition (and (holding ?b1) (clear ?b2) (< (size ?b1) (size ?b2)))}|
    \end{minted}
    \end{tcolorbox}
    \vspace*{-0.6em}
    \caption{Automatic modification of the generated PDDL domain (bottom) to incorporate additional requirements (top).}
    \label{fig:adaptability-diff}
\end{figure}

A similar process was used to add a \texttt{battery\_level} fluent and modify the effects of \texttt{move} and \texttt{paint} actions to consume battery. The natural language goal prompt specified the precise cost of each action (e.g., ``each move action costs 5 battery units'') and the final goal condition (e.g., ``robot must have at least 20 battery units left''). The agent was then expected to query the user for the initial battery levels.

The results demonstrate TAPAS's ability to autonomously refine domain models, a capability beyond traditional planners. The lower success rate in \textit{floortile} was primarily due to the agent assuming an initial battery level instead of querying the user, highlighting the ongoing challenge of preventing agent assumptions.


\begin{figure}[b]
    \begin{tcolorbox}[fontupper=\small, colframe=blue!50!black, colback=blue!2, boxrule=0.5pt, arc=2pt, left=3pt, right=3pt, top=1.5pt, bottom=1.5pt]
    \textit{``Save the following to memory: for problems involving the fridge, append a goal to close the fridge, even if not explicitly stated.''}
\end{tcolorbox}
    \vspace*{-0.6em}
    \caption{Explicit user feedback for the procedural memory.}
    \label{fig:user-instruction}
\end{figure}

\subsection{Evaluation in VirtualHome Simulation}

\begin{figure*}[t]
\centering
    \subfloat[Open fridge]{%
        \includegraphics[width=0.24\textwidth]{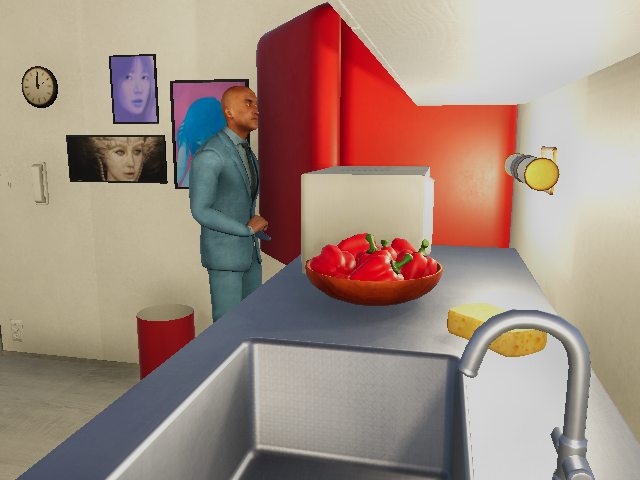}%
    }
    \hfill
    \subfloat[Grab salmon]{%
        \includegraphics[width=0.24\textwidth]{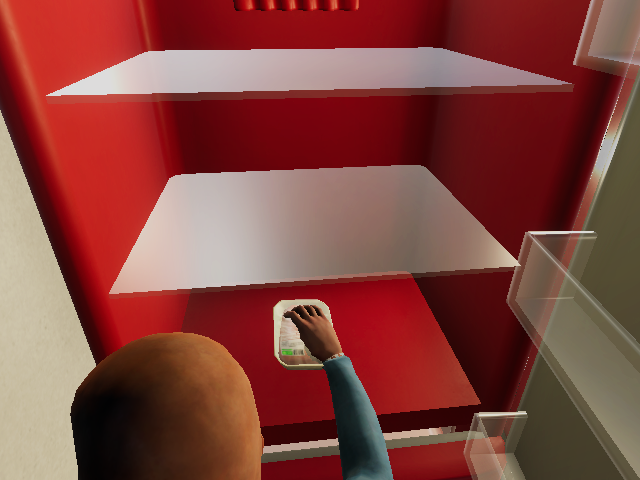}%
    }
    \hfill
    \subfloat[Heat up]{%
        \includegraphics[width=0.24\textwidth]{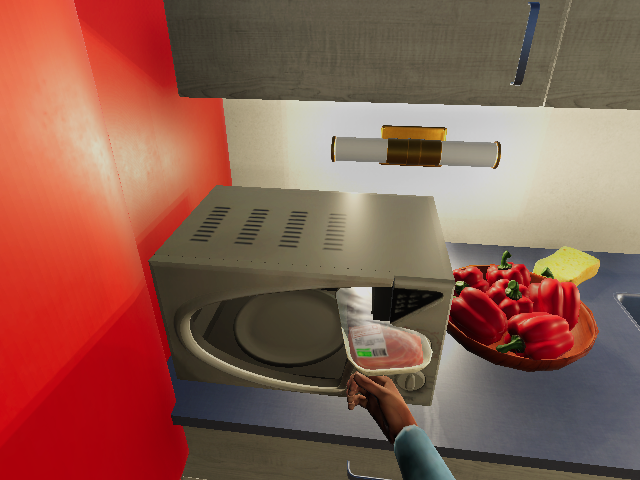}%
    }
    \hfill
    \subfloat[Place salmon on kitchen table]{%
        \includegraphics[width=0.24\textwidth]{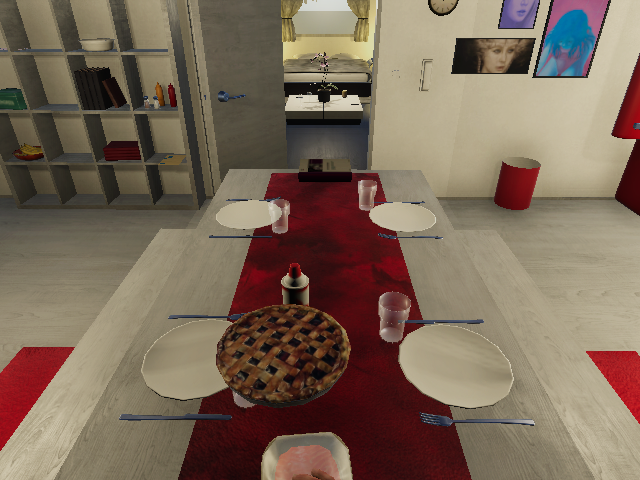}%
    }
\caption{Plan execution in VirtualHome: The humanoid agent sequentially interacts with objects to complete the task. (a) Opening fridge, (b) salmon retrieval, (c) heating, and (d) placement on the kitchen table.}
\label{fig:virtualhome_execution}
\end{figure*}

In addition to the benchmarking study, experiments were conducted within the VirtualHome simulation.
We provide this evaluation to showcase the application of the entire TAPAS framework in a complex planning and execution pipeline.
VirtualHome provides a 3D household simulation where humanoid agents interact with objects to execute tasks from language instructions.

\paragraph{Experimental Setup}
The task is to achieve two goals: placing \texttt{pie} on \texttt{kitchentable} and warming and placing \texttt{salmon} on the same table. TAPAS received task descriptions and initial environment states in natural language. The system generated structured plans, translated them into executable instructions, and sequentially executed each step within the simulation. The \textit{plan executor} interacted with the environment using predefined skills, receiving first-person visual feedback and textual responses. In this simulated environment, information like object states and locations is obtained directly from the simulation state, which is then converted into the textual and visual observations provided to the executor, simulating the output of a real-world perception stack. Execution continued until all plan steps were completed, and goal satisfaction was verified.

\paragraph{Results on Execution}
TAPAS successfully generated and executed structured plans to achieve the specified goals. The humanoid agent interacted with the fridge, retrieved the salmon, used the microwave for heating, and placed it on the kitchen table. The system demonstrated effective translation of high-level goals into executable steps, showcasing robust task planning and execution.

To evaluate TAPAS's procedural memory, the goal state generator received an instruction mimicking user feedback as shown in Figure~\ref{fig:user-instruction}
Upon encountering a similar task after context clearance, TAPAS successfully recalled and applied the stored instruction. The planner autonomously added the goal condition: \mintinline[fontsize=\small, style=manni]{lisp}{(Not (is_open fridge_305))}. This ensured fridge closure despite its absence from the new task description. This demonstrates TAPAS's potential for long-term adaptation across tasks, though further evaluation of consistency and scalability is necessary.

\paragraph{Plan Execution in VirtualHome}
Figure~\ref{fig:virtualhome_execution} illustrates key plan execution steps. Snapshots depict the humanoid agent retrieving, heating, and placing food items. This demonstrates TAPAS's ability to translate abstract goals into executable steps while adapting to dynamic visual feedback.


\section{Discussion}

TAPAS's core contribution is its ability to autonomously generate and adaptively update domain models, initial states, and goal specifications, facilitated by collaborative LLM agents. This adaptive mechanism distinguishes it from static, predefined planning approaches, demonstrated by its successful adaptation to novel attributes in \textit{blocksworld} and execution in VirtualHome.

The performance of TAPAS is intrinsically linked to the capabilities of the underlying LLM, as confirmed by our model ablation studies. Overall, Claude 3.7 Sonnet and GPT-4o achieved the highest performance with a clear margin over the other investigated LLMs. This underscores the importance of selecting appropriate models and investigating their generalizability.

Mitigating LLM hallucinations and errors remains crucial for robust real-world deployment. While self-reflection and refinement mechanisms address certain errors, deeper biases and complex hallucination, observed particularly with smaller models, remain a challenge.
While a detailed analysis of this behavior is beyond the scope of this work, this problem could potentially be mitigated by the inclusion of in-domain examples.

In application scenarios, error detection during plan execution relies on the Validator Agent's role in identifying discrepancies between executed actions and the expected state. While the system can attempt corrective feedback loops, irrecoverable errors such as fundamental task misunderstandings currently lead to user notification.

Future research can build upon the presented framework to enhance its robustness and expand its capabilities. For instance, rigorous ablation studies could quantify the contribution of individual components to inform the design of more efficient neuro-symbolic architectures. The system's generalization could be further tested through evaluation against larger benchmarks and in less-structured, dynamic environments. Such testing would identify critical failure modes and guide improvements in planner robustness, including strategies to achieve strong performance with smaller, resource-efficient LLMs.

A key long-term objective is to enhance operational autonomy by developing sophisticated closed-loop execution cycles. Enabling the system to recover from errors by re-engaging the planner is a critical step towards reliable real-world deployment.


\section{Conclusion}

TAPAS introduces a significant advancement in adaptable planning systems for open-world environments. By dynamically adapting domain models and effectively handling novel constraints, TAPAS achieved an 88.42\% success rate across diverse benchmark planning domains and demonstrated successful execution in a simulated real-world setting. These results highlight the potential of integrating LLM-driven adaptability with the structured reasoning of symbolic planning.

Future research will focus on incorporating hierarchical and temporal reasoning, expanding TAPAS's capabilities to a broader range of environments, and integrating human-in-the-loop feedback. TAPAS provides a robust foundation for developing adaptable and trustworthy planning agents capable of navigating the complexities of real-world tasks.

The code will be available shortly at \url{https://sites.google.com/view/adaptive-llm-planning}.


\addtolength{\textheight}{-0cm} 


\section*{ACKNOWLEDGMENT}

We thank Patrick Kesper and Dragan Milchevski for support in setting up endpoints on AWS and Azure, respectively.


\bibliographystyle{IEEEtranN}
\small
\bibliography{IEEEabrv,references}

\end{document}